\documentclass[runningheads]{llncs}

\usepackage{eccv}

\usepackage{eccvabbrv}
\usepackage[table]{xcolor}
\usepackage{graphicx}
\usepackage{booktabs}

\usepackage[accsupp]{axessibility}  

\usepackage{hyperref}

\usepackage{orcidlink}

\begin{document}

\title{Seeing the Unseen: Semantic-in-Gaussian for Sparse-View 3D Generalization}

\titlerunning{SeeU}

\author{Zeyang Bai\inst{1}\orcidlink{0009-0003-8776-6980} \and
Yunpeng Wang\inst{2} \and
Yunbiao Wang\inst{1}\textsuperscript{*}\orcidlink{0000-0001-9648-6073} \and
Jun Xiao\inst{1}\textsuperscript{*}\orcidlink{0000-0002-1799-3948}
}

\authorrunning{Z.~Bai et al.}

\institute{School of Artificial Intelligence,
University of Chinese Academy of Sciences, China \and
Global Institute of Future Technology,
Shanghai Jiao Tong University, China}

\maketitle
\begingroup
\renewcommand{\thefootnote}{\fnsymbol{footnote}}
\footnotetext[1]{Corresponding authors.}
\endgroup

\begin{figure}[!th]
  \centering
  \vspace{-24pt}
  \includegraphics[width=\linewidth]{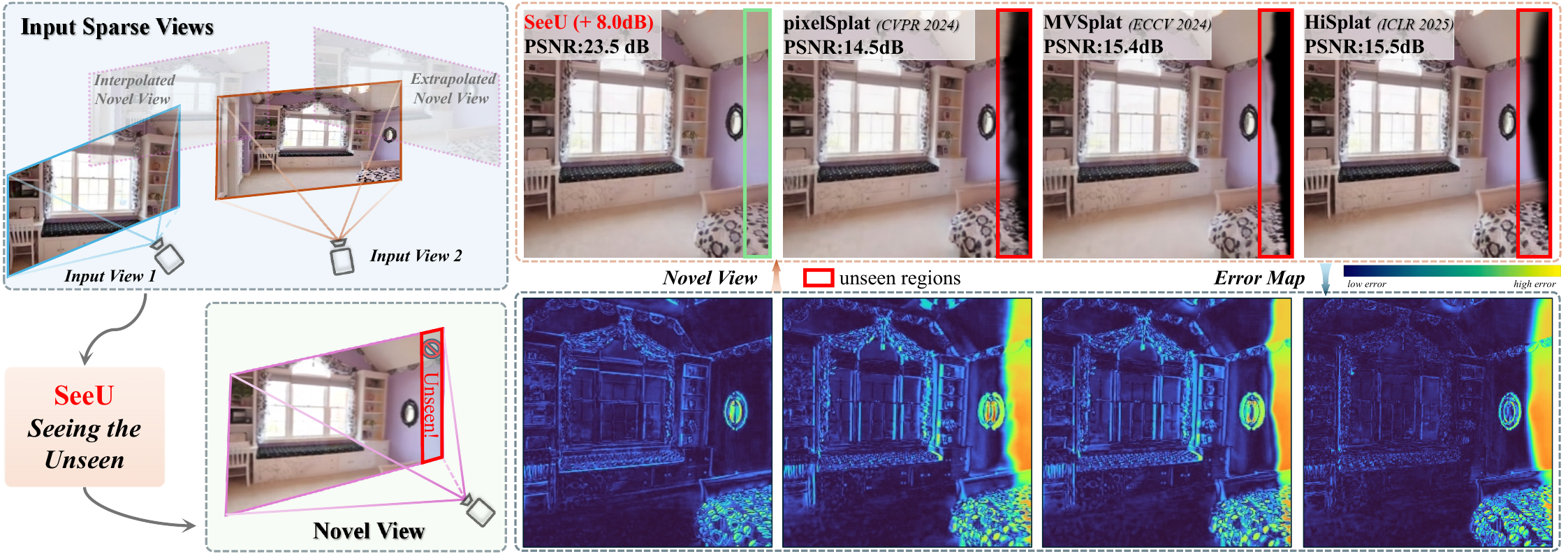}
  \vspace{-12pt}
  \caption{\textbf{Comparison of SeeU with sparse-view G-3DGS baselines.}
  Previous methods are pixel-aligned and depth-based, often failing to recover geometry in ambiguous or occluded regions (e.g., unseen surfaces).
  In contrast, SeeU leverages semantic embeddings to guide the Conditional Gaussian Transformer, which refines 3D Gaussians to enable high-fidelity reconstruction in uncertain regions.
  Residual maps against the ground-truth novel view are shown in the bottom row.}
  \vspace{-32pt}
  \label{fig:teaser}
\end{figure}

\begin{abstract}
Generalizable 3D Gaussian Splatting (G-3DGS) has emerged as a promising approach for novel view synthesis under sparse-view settings.
However, existing frameworks remain restricted by pixel-aligned Gaussian estimation, which struggles in partially observed or occluded regions and often leads to incomplete surfaces or structural collapse. 
To address these challenges, we propose \textbf{SeeU} (\textbf{See}ing the \textbf{U}nseen), a novel G-3DGS framework.
We frame its core design as \emph{Semantic-in-Gaussian}: semantic-conditioned refinement in Gaussian space.
Specifically, we introduce a Cross-view Entropy-Aware (CEA) module that aggregates multi-view semantic and geometric cues into compact embeddings.
These embeddings guide the Conditional Gaussian Transformer, which applies residual updates to coarse Gaussians, helping recover under-constrained regions of partially observed structures while preserving surface consistency.
Comprehensive experiments on multiple benchmarks demonstrate that SeeU consistently improves rendering quality and structural completeness while retaining efficient feed-forward inference.
Especially under challenging extrapolation settings, SeeU achieves an average improvement of \textbf{2.44 dB} in PSNR compared to recent SOTA G-3DGS methods.  
\end{abstract}

\section{Introduction}
\label{sec:intro}
3D reconstruction is a cornerstone of computer vision, powering applications such as autonomous driving, virtual reality, and augmented reality. 
Recently, 3D Gaussian Splatting (3DGS)~\cite{tog20233dgs} has emerged as a powerful paradigm for scene representation and novel view synthesis (NVS)~\cite{siggraph2001nvs}. 
By modeling a scene as a mixture of 3D Gaussians and leveraging differentiable rasterization, 3DGS enables high-fidelity and real-time rendering from dense multi-view images. 
However, conventional 3DGS methods typically rely on per-scene optimization with dense input views, which limits scalability and requires costly data acquisition.
To alleviate these constraints, generalizable 3DGS (G-3DGS) has been developed to reconstruct scenes from only a few input views.
These approaches employ pre-trained feed-forward models~\cite{MVSplat-2025, PixelSplat-2024, aaai2025transplat, GraphSplat-2025, latentsplat, HiSplat-2024, eFreeSplat-2024, cvpr2025depthsplat} that encode scene priors from large-scale datasets, enabling rapid inference without scene-specific optimization.

Despite this progress, existing G-3DGS frameworks remain fundamentally restricted by their strong reliance on pixel-aligned unprojection~\cite{PixelSplat-2024, MVSplat-2025}. 
In these frameworks, estimated depth anchors Gaussian centers to input pixels, causing depth errors to propagate directly into the reconstructed geometry.
However, under sparse-view conditions, depth estimation often suffers from occlusions, weak textures, and limited viewpoint overlap.
Consequently, pixel-aligned estimation provides insufficient support for occluded or weakly observed parts of partially visible structures, often producing `black holes' or collapsed geometry in the rendered novel views (Fig.~\ref{fig:teaser}).

While previous G-3DGS works attempted to address uncertainty through depth regularization or feature fusion~\cite{aaai2025transplat, cvpr2025depthsplat}, they remain inherently constrained by pixel-level priors. 
Motivated by semantic conditioning in text-to-3D generation~\cite{Text-to-3D-2025, Large-Vocabulary-3D-Diffusion-2024}, we investigate whether semantic priors can compensate for pixel-level uncertainty.
Building on this intuition, we introduce \textbf{SeeU}, a feed-forward G-3DGS framework that \textbf{shifts reconstruction from direct pixel-space estimation to semantic-guided refinement in Gaussian space}.
Specifically, SeeU first constructs coarse pixel-aligned latent Gaussians from sparse input views.
However, a central challenge is obtaining an effective semantic condition because NVS provides no explicit text prompts.
Pseudo-captioning methods such as BLIP~\cite{icml2023blip} offer global semantic cues but often overlook fine-grained structures~\cite{cvpr2024ecodepth}.
We therefore design a \textbf{Cross-view Entropy-Aware (CEA)} module that aggregates multi-view semantic cues and uses depth-distribution entropy to emphasize weakly constrained regions.
Conditioned on the resulting embeddings, a \textbf{Conditional Gaussian Transformer} predicts residual updates to the coarse Gaussians.
This refinement supports the recovery of missing surfaces in semantically anchored and partially observed structures while preserving consistency with the input views.
By applying a residual refinement within a feed-forward pipeline, SeeU retains efficient inference and reduces its dependence on precise depth alignment.

The main contributions are summarized as follows:
\vspace{-0.5em}
\begin{itemize}
\item We introduce \textbf{SeeU}, a framework for \textbf{semantic}-conditioned refinement \textbf{in Gaussian} space. SeeU employs a Conditional Gaussian Transformer to perform residual refinement on coarse Gaussians, reducing its dependence on pixel-aligned depth estimation.
\item We propose a Cross-view Entropy-Aware module that combines multi-view semantic cues with depth-distribution entropy, providing structure-aware conditioning for weakly constrained regions.
\item We evaluate SeeU across interpolation, extrapolation, and zero-shot cross-dataset settings, consistently improving rendering quality. On extrapolation of RealEstate10K~\cite{RealEstate10K-2018}, SeeU improves PSNR by \textbf{+2.32 dB} over recent G-3DGS SOTA method HiSplat~\cite{HiSplat-2024}.
\end{itemize}

\section{Related Work}\label{sec:relate}
\subsection{Novel View Synthesis}
Novel view synthesis (NVS) aims to render photo-realistic images from novel viewpoints using only a limited set of input images \cite{siggraph2001nvs}. 
Neural Radiance Fields (NeRF) \cite{NeRF2020, PixelNeRF2021, D-NeRF2021,Mip-NeRF2021, Mip-NeRF360-2022, cvpr2024murf} models scenes as continuous volumetric radiance fields parameterized by neural networks. 
While NeRF-based methods have yielded impressive results in dense multi-view settings, they typically suffer from slow training times, high memory usage, and suboptimal performance under sparse viewpoints due to their heavy reliance on per-ray MLP evaluations. 
In contrast, 3D Gaussian Splatting (3DGS) \cite{tog20233dgs, Mip-Splatting-2024,Deformable-3DGS-2024} introduces an explicit scene representation by modeling surfaces with anisotropic 3D Gaussian primitives. 
Each Gaussian is described by its position, covariance, color, and opacity, which can be differentiably rendered via a forward projection to the image plane. 
This explicit design significantly accelerates the rendering process compared to vanilla NeRF pipelines, yet many existing 3DGS approaches still assume relatively dense coverage of views for accurate geometry and appearance reconstruction. 
Consequently, their performance deteriorates for extremely sparse inputs, where geometric ambiguity and insufficient texture cues become major challenges. 
\begin{figure*}[!ht]
    \centering
    \includegraphics[width=\linewidth]{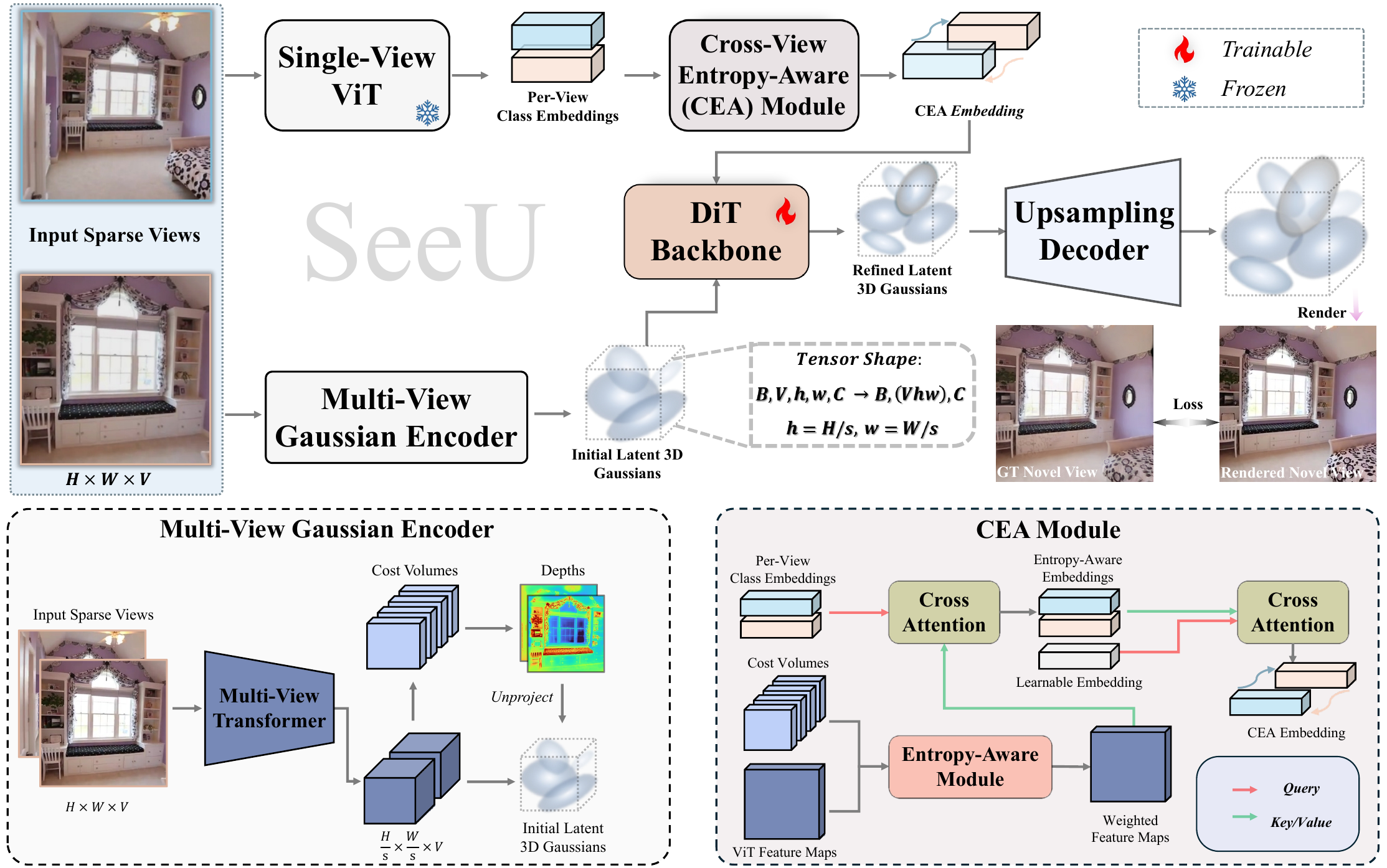}
    \caption{
\textbf{Overview of SeeU.} 
Given sparse input views, our framework first constructs an initial set of latent 3D Gaussians through the multi-view Gaussian encoder.
In parallel, a frozen single-view ViT extracts per-view class embeddings, which are fused by the proposed Cross-view Entropy-Aware (CEA) module to produce a unified multi-view semantic embedding.
This embedding guides the DiT-based Conditional Gaussian Transformer to perform residual refinement on the latent Gaussians, followed by an upsampling decoder that restores them to the original spatial resolution.
The refined 3D Gaussians are rendered to synthesize novel views, and the entire model is optimized end-to-end using photometric loss.
}
    \label{fig:overview}
\end{figure*}

\subsection{Sparse-View Generalizable 3DGS}
Sparse-View generalizable 3DGS methods focus on learning a feed-forward model capable of handling unseen scenes without per-scene re-optimization~\cite{GraphSplat-2025}. 
PixelSplat \cite{PixelSplat-2024} predict 3D Gaussian parameters from sparse multi-view inputs, leveraging an epipolar transformer for depth estimation. 
MVSplat \cite{MVSplat-2025} relies on cost-volume construction via plane sweeping to infer depth distributions.
TranSplat \cite{aaai2025transplat} introduces a transformer-based architecture with depth-aware deformable matching for coarse-to-fine refinement. 
HiSplat \cite{HiSplat-2024} integrates hierarchical Gaussian features, leveraging iterative Gaussian alignment.
eFreeSplat~\cite{eFreeSplat-2024} eliminates epipolar priors by leveraging cross-view completion.
DepthSplat~\cite{cvpr2025depthsplat} bridges Gaussian splatting and depth estimation by leveraging pre-trained monocular depth features to enhance multi-view depth prediction. 
Despite these diverse approaches, most still rely heavily on estimated depth maps, which can become noisy or unreliable under sparse-view conditions. 
They also often utilize pixel-aligned Gaussian estimation, causing difficulties in recovering fine details or resolving ambiguities in unseen regions. 

\subsection{Semantic-Conditioned Gaussian Methods}

Conditional 3D generation methods use text or image cues to guide content creation.
Text-conditioned methods generate point clouds or other 3D representations from language prompts~\cite{Point-E-2022,Large-Vocabulary-3D-Diffusion-2024}, while image-conditioned methods reconstruct 3D assets from single-view or generated multi-view observations~\cite{iclr2024lrm,Instant3D-2024,Large-Multiview-Gaussian-Model-2025,Multiview-Diffusion-for-3D-Generation-2024}.
Recent approaches further generate Gaussian representations directly using diffusion-based models~\cite{iclr2025diffsplat,cvpr2025prometheus}.
However, these methods primarily target 3D content generation, whereas sparse-view scene reconstruction requires fidelity to observed views and multi-view geometric consistency.
SeeU therefore does not generate a scene from noise or employ a diffusion objective.
Instead, it uses cross-view semantic embeddings from CEA to condition the Conditional Gaussian Transformer, which performs residual refinement on coarse pixel-aligned Gaussians.
This design provides semantic compensation for under-constrained regions while preserving structural alignment with the input views.

\section{Methodology}\label{sec:method}
Following the G-3DGS framework~\cite{PixelSplat-2024,MVSplat-2025,aaai2025transplat}, the input consists of $V$ sparse-view images $\mathcal{I} = \{I^1, I^2, \dots, I^V\}$, where each image $I^i \in \mathbb{R}^{H \times W \times 3}$ is accompanied by its camera projection matrix derived from intrinsic and extrinsic parameters.  
The goal is to reconstruct the underlying 3D scene as a set of Gaussian primitives 
\(\Theta = \{G_j\}_{j=1}^N\), 
where each primitive \(G_j\) is parameterized by its center \(\boldsymbol{\mu}_j\), opacity \(\alpha_j\), covariance \(\boldsymbol{\Sigma}_j\), and color \(\boldsymbol{c}_j\).  
The number of Gaussians is typically set to \(N = H \times W \times V\), 
corresponding to the input resolution and number of views.  
These primitives are subsequently rendered into novel views through differentiable Gaussian splatting~\cite{tog20233dgs}.

As illustrated in Figure~\ref{fig:overview}, SeeU performs semantic-conditioned refinement directly in Gaussian space.
The pipeline comprises four stages: Gaussian initialization, Cross-view Entropy-Aware (CEA) conditioning, residual refinement, and Gaussian decoding.
First, a multi-view Gaussian encoder extracts cross-view features and constructs coarse pixel-aligned latent Gaussians from the sparse input views.
The CEA module then combines semantic features from a frozen ViT with matching uncertainty derived from the cost volumes, producing a cross-view conditioning embedding.
Guided by this embedding, the DiT-based Conditional Gaussian Transformer predicts residual updates to refine the coarse latent Gaussians.
Finally, an upsampling decoder maps the refined representation to full-resolution Gaussian centers, opacity, covariance, and color for novel-view rendering.

\subsection{Gaussian Initialization}
A reliable coarse initialization anchors the subsequent residual refinement to the observed views.
We use a multi-view Gaussian encoder to construct latent 3D Gaussians directly from sparse input views.
By aggregating multi-view features and estimating coarse scene geometry, the encoder produces a structured and pixel-aligned Gaussian representation.
These initialized Gaussians serve as the input to the Conditional Gaussian Transformer, which refines their parameters under CEA conditioning.
\subsubsection{Latent Feature Extraction}\label{sec:latent-feature}
To control the computational cost of subsequent Gaussian refinement, we extract multi-view features at a reduced spatial resolution.
Each input image $I^i$ is first processed by a shallow ResNet~\cite{cvpr2016resnet} to produce an $s$-fold downsampled feature map.
A multi-view Swin Transformer~\cite{iccv2021swin} with cross-view attention then aggregates information across the input views, producing features $\boldsymbol{F}^i \in \mathbb{R}^{\frac{H}{s} \times \frac{W}{s} \times C}$, where $C$ denotes the feature dimension.
This latent representation preserves cross-view interactions while reducing the sequence length processed during Gaussian refinement.
We use the resulting features to initialize the latent Gaussian parameters.

\subsubsection{Coarse Matching}
To initialize coarse Gaussian geometry and estimate matching uncertainty for CEA conditioning, we construct cost volumes with plane sweeping~\cite{pami2023uniflow,eccv2018mvsnet} to model multi-view feature correspondences.
Specifically, for each view $i$, we uniformly sample $D$ depth candidates $\{d_m\}_{m=1}^D$ in the inverse depth domain between near and far planes. 
Features from other views ($\boldsymbol{F}^j, j \neq i$) are warped to view $i$ using camera parameters and depth candidate $d_m$, producing $D$ warped features $\{\boldsymbol{F}_{d_m}^{j \rightarrow i}\}_{m=1}^D$.
The correlation $\boldsymbol{C}_{d_m}^i$ between $\boldsymbol{F}^i$ and $\boldsymbol{F}_{d_m}^{j \rightarrow i}$ is computed with the dot-product operation:
\begin{equation}
\boldsymbol{C} _{d_m}^i = \frac{F_i \cdot F^{j \to i}_{d_m}}{\sqrt{C}}, 
\quad m = 1, 2, \ldots, D.
\end{equation}
We average $\boldsymbol{C}_{d_m}^i$ across all other views to form the cost volume $\mathbf{C}^i \in \mathbb{R}^{\frac{H}{s} \times \frac{W}{s} \times D}$.
Subsequently, we apply the softmax operation to compute per-view depth map: 
\begin{equation}
\boldsymbol{Z}^i = \text{softmax}(\mathbf{C}^i) \cdot \boldsymbol{\Lambda}, 
\end{equation}
where $\boldsymbol{\Lambda} = [d_1, d_2, \dots, d_D]$ are the depth candidates. 
The coarse depth map $\boldsymbol{Z}^i \in \mathbb{R}^{\frac{H}{s} \times\frac{W}{s}}$ is then unprojected to form preliminary Gaussian centers $\boldsymbol{\mu}$ using the camera parameters, and other Gaussian parameters are predicted by additional lightweight heads from feature maps.
This ensures that the initial positions are geometrically consistent with the input views.
The constructed cost volumes $\mathbf{C}^i$ are further retained as conditional inputs to the CEA module, providing pixel-wise structural cues about the scene.

\subsection{Conditional Gaussian Transformer}
Coarse pixel-aligned Gaussians remain susceptible to depth errors in weakly constrained regions.
To correct these errors without discarding geometry anchored by the input views, we introduce a DiT-based Conditional Gaussian Transformer for residual refinement in Gaussian space.
Conditioned on CEA embeddings, the module predicts residual updates to the latent Gaussian parameters.

\subsubsection{Cross-view Entropy-Aware Embedding}
Global conditioning signals, such as BLIP-derived text features~\cite{icml2023blip}, capture scene-level semantics but may underrepresent fine-grained spatial structures, providing weaker guidance near object boundaries~\cite{cvpr2024ecodepth}.
When extracted independently from each input view, these signals also retain overlapping content without explicitly consolidating cross-view evidence.
We therefore introduce a Cross-view Entropy-Aware (CEA) module that uses matching entropy to emphasize weakly constrained regions and aggregates per-view semantics into a compact multi-view embedding.

We first propose an entropy-aware module that leverages the cost volume $\mathbf{C}^i$ to compute the matching entropy $H^i$, thereby identifying weakly constrained regions (e.g., occluded, weakly observed, or textureless areas).
A frozen single-view ViT encoder pretrained with DINOv3~\cite{simeoni2025dinov3} extracts a class embedding $\mathbf{E}^i_{\mathrm{CLS}}$ and a spatial feature map $\mathbf{F}^i$ from each view.
For each pixel $p$, we obtain a depth posterior by applying softmax along the depth dimension of the cost volume:
\begin{equation}
    P^i(d \mid p) = \mathrm{softmax}_d\big(\mathbf{C}^i(p,d)\big).
\end{equation}
We then compute the matching entropy as
\begin{equation}
    H^i(p) = -\sum_{d \in \Lambda} P^i(d \mid p)\log P^i(d \mid p).
\end{equation}
Higher entropy indicates greater ambiguity in multi-view matching.
We normalize the entropy by its maximum value over $D$ depth candidates:
\begin{equation}
    w^i(p) = \frac{H^i(p)}{\log D},
    \qquad w^i(p) \in [0,1].
\end{equation}
The resulting weight emphasizes weakly constrained locations in the spatial feature map:
\begin{equation}
    \tilde{\mathbf{F}}^i(p) = w^i(p)\mathbf{F}^i(p).
\end{equation}
We project the class embedding into a query and the entropy-weighted spatial features into keys and values:
\begin{equation}
    \mathbf{Q}^i = \mathbf{E}^i_{\mathrm{CLS}}\mathbf{W}_Q,
    \quad \mathbf{K}^i = \tilde{\mathbf{F}}^i \mathbf{W}_K,
    \quad \mathbf{V}^i = \tilde{\mathbf{F}}^i \mathbf{W}_V.
\end{equation}
Cross-attention then incorporates uncertainty-weighted spatial cues into the per-view semantic embedding:
\begin{equation}
    \tilde{\mathbf{E}}^i_{\mathrm{CLS}}
    = \mathrm{CrossAttn}\!\left(\mathbf{Q}^i,\mathbf{K}^i,\mathbf{V}^i\right).
\end{equation}

To aggregate complementary evidence across views while compressing overlapping content, we employ cross-view Perceiver-style attention with a set of learnable latent queries $\mathbf{Q}_{\ell}$.
We concatenate the refined per-view embeddings $\tilde{\mathbf{E}}^i_{\mathrm{CLS}}$ and project them into keys $\mathbf{K}_{\mathrm{mv}}$ and values $\mathbf{V}_{\mathrm{mv}}$.
The CEA embedding is then computed as
\begin{equation}
    \mathbf{E}_{\mathrm{CEA}} = \mathrm{CrossAttn}\!\big(\mathbf{Q}_{\ell},\, \mathbf{K}_{\mathrm{mv}},\, \mathbf{V}_{\mathrm{mv}}\big).
\end{equation}
$\mathbf{E}_{\mathrm{CEA}}$ summarizes uncertainty-aware semantics across views.
It conditions the residual refinement performed by the Conditional Gaussian Transformer.

\subsubsection{Gaussian-Structured Representation}
Although Gaussian primitives conceptually form a set, our initialization preserves a pixel-to-Gaussian correspondence.
We therefore organize the latent parameters as $\Theta_l \in \mathbb{R}^{B \times V \times h \times w \times C}$, where $B$ is the batch size, $V$ is the number of views, $h \times w = \frac{H}{s} \times \frac{W}{s}$ is the latent spatial resolution, and $C$ is the Gaussian parameter dimension.
Applying a convolutional architecture such as UNet~\cite{miccai2015unet} would impose fixed local grid neighborhoods on this tensor, although nearby image locations need not correspond to nearby Gaussians in 3D.
We instead use a DiT-based Transformer module to model interactions among Gaussian tokens.
Specifically, we flatten the structured tensor into a token sequence:
\begin{equation}
    \mathbf{T}_l = \texttt{rearrange}(\Theta_l,\ \texttt{B V h w C} \rightarrow \texttt{B (Vhw) C}),
\end{equation}
where $\mathbf{T}_l \in \mathbb{R}^{B \times (Vhw) \times C}$ is processed by the DiT architecture~\cite{iccv2023dit}.
Rather than regressing the refined Gaussians directly, the module predicts a residual update to the coarse parameters:
\begin{equation}
\hat{\Theta}_l = \Theta_l + f_\theta(\Theta_l, \mathbf{E}_{\mathrm{CEA}}),
\end{equation}
where $f_\theta$ is the DiT-based predictor conditioned on the CEA embedding $\mathbf{E}_{\mathrm{CEA}}$.
This residual formulation corrects errors in weakly constrained regions while preserving geometry that is already well anchored by the input views.

\begin{table}[t]
\centering
\caption{\textbf{Comparison of interpolated NVS.}
We evaluate performance on the RealEstate10K and ACID datasets by rendering three novel interpolation views from two reference viewpoints, averaging across all scenes.
The dataset’s training and testing split follows the identical protocol established by pixelSplat.
Note that 3DGS-based methods render extremely fast ($\sim$ 500FPS).
}

\resizebox{1.\linewidth}{!}{
\begin{tabular}{lccccccc}
\hline
& \multicolumn{3}{c}{\textbf{RealEstate10K}} & \multicolumn{3}{c}{\textbf{ACID}} & \textbf{Inference Time} \\
Method & PSNR$\uparrow$ & SSIM$\uparrow$ & LPIPS$\downarrow$ & PSNR$\uparrow$ & SSIM$\uparrow$ & LPIPS$\downarrow$ & (s) \\
\hline
pixelNeRF~\cite{PixelNeRF2021} & 20.43 & 0.589 & 0.550 & 20.97 & 0.547 & 0.533 & 5.299 \\
MuRF~\cite{cvpr2024murf} & 26.10 & 0.858 & 0.143 & 28.09 & 0.841 & 0.155 & 0.186 \\
\hline
pixelSplat~\cite{PixelSplat-2024}  & 25.89 & 0.858 & 0.142 & 28.14 & 0.839 & 0.150 & 0.104 \\
MVSplat~\cite{MVSplat-2025}        & 26.39 &0.869 &0.128 & 28.25 &0.843 & 0.144 &\textbf{0.044} \\
eFreeSplat~\cite{eFreeSplat-2024}  & 26.45 & 0.865 & 0.126 & 28.30 &\cellcolor{yellow!20}0.851 & \cellcolor{yellow!20}0.140 & 0.061 \\
TranSplat~\cite{aaai2025transplat} & \cellcolor{yellow!20}26.69 & \cellcolor{yellow!20}0.875 & \cellcolor{yellow!20}0.125 & \cellcolor{yellow!20}28.35 &0.845 & 0.143 & 0.087 \\
HiSplat~\cite{HiSplat-2024}        & \cellcolor{orange!20}27.21 & \cellcolor{orange!20}0.881 & \cellcolor{orange!20}0.117 & \cellcolor{orange!20}28.75 & \cellcolor{orange!20}0.853 & \cellcolor{red!20}\textbf{0.133} & 0.510 \\
\textbf{SeeU (Ours)}        & \cellcolor{red!20}\textbf{27.56} & \cellcolor{red!20}\textbf{0.888} & \cellcolor{red!20}\textbf{0.114} & \cellcolor{red!20}\textbf{28.77} & \cellcolor{red!20}\textbf{0.855} & \cellcolor{red!20}\textbf{0.133} & 0.089 \\
\hline
\end{tabular}}
    \vspace{-1.2em}

\label{tab:inter-compare}
\end{table}

\begin{figure*}[!t]
    \centering
    \includegraphics[width=\linewidth]{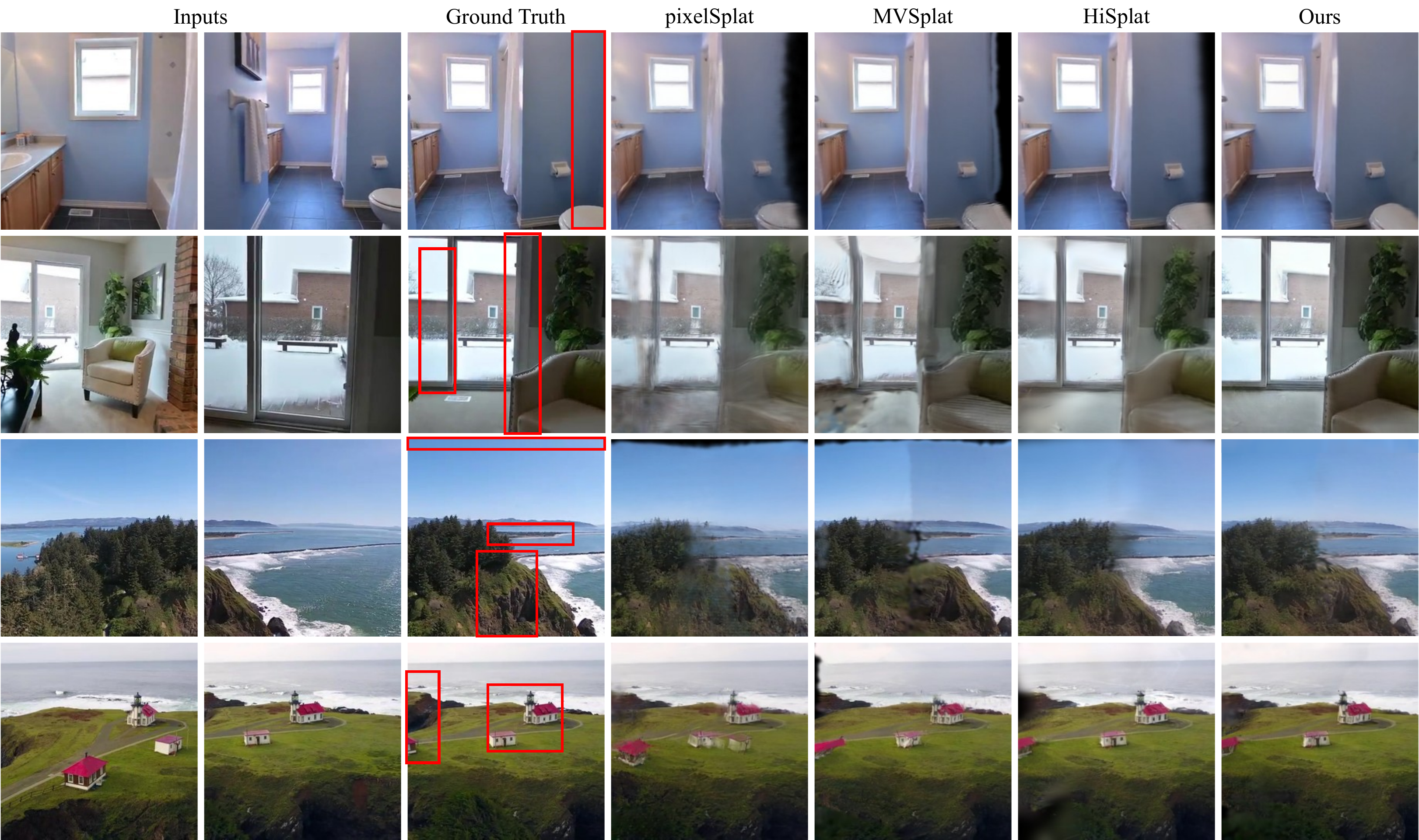}
    \caption{\textbf{Qualitative comparison of interpolated NVS.}
    The first two columns show sparse input views, while the third column presents the ground truth for the target interpolated view between them. 
    SeeU better reconstructs fine details and occluded regions (highlighted in red) in both indoor (RealEstate10K, top two rows) and outdoor (ACID, bottom two rows) settings.
    }
    \vspace{-1.5em}
    \label{fig:inter-compare}
\end{figure*}

\subsection{Rendering and Training Loss}
The upsampling decoder produces the full-resolution Gaussian set $\Theta$, which is rendered from target viewpoints using differentiable Gaussian rasterization~\cite{tog20233dgs}.
We optimize SeeU end-to-end through image-space supervision.
The training objective combines a pixel-wise $\ell_2$ reconstruction term with an LPIPS perceptual term~\cite{cvpr2018lpips}:
\begin{equation}
\mathcal{L}_{\text{photo}}
=
\|\mathcal{R}(\Theta)-\mathcal{I}_{\text{gt}}\|_2^2
+
0.05 \cdot \mathrm{LPIPS}\!\left(
\mathcal{R}(\Theta),\mathcal{I}_{\text{gt}}
\right),
\end{equation}
where $\mathcal{R}(\Theta)$ denotes the target-view image rendered from $\Theta$.
Following prior G-3DGS methods~\cite{PixelSplat-2024,MVSplat-2025}, we set the LPIPS weight to $0.05$.

\section{Experiments and Discussions}\label{sec:exp}
\subsection{Experimental Settings}
For within-dataset experiments, we train and evaluate SeeU on two large-scale datasets, RealEstate10K~\cite{RealEstate10K-2018} and ACID~\cite{ACID-2021}.
RealEstate10K contains 67,477 training scenes and 7,289 test scenes, while ACID contains 11,075 training scenes and 1,972 test scenes.
Both datasets provide per-frame camera intrinsics and extrinsics estimated through Structure-from-Motion (SfM)~\cite{Structure-from-Motion-2016}.
Following prior works~\cite{PixelSplat-2024,MVSplat-2025}, we use two context images and render three target views for each test scene.
We evaluate both interpolated and extrapolated NVS, with extrapolated target views lying outside the reference-view range.
For extrapolation training, we follow the pixelSplat curriculum~\cite{PixelSplat-2024} and increase the target-view sampling interval to 45 frames before and after the reference views.
We report PSNR, SSIM~\cite{SSIM-2004}, and LPIPS~\cite{LPIPS-2018} for rendering quality.
For zero-shot cross-dataset evaluation, the model trained on RealEstate10K is evaluated on 16 DTU validation scenes~\cite{DTU-dataset-2014}, with four novel target views per scene and no fine-tuning, following the MVSplat protocol~\cite{MVSplat-2025}.
On DTU, we additionally report depth RMSE and Overall Chamfer to assess geometric accuracy.
Implementation details, ACID cross-dataset results, and evaluations with more input views are provided in the Appendix.

\subsection{Main Results}

\subsubsection{Interpolated Novel View Synthesis}
Table~\ref{tab:inter-compare} shows that SeeU achieves better performance on both RealEstate10K and ACID.
Figure~\ref{fig:inter-compare} shows that, compared with pixel-aligned G-3DGS baselines, SeeU better preserves fine structures and reduces artifacts in occluded regions through CEA-conditioned residual refinement.
The full framework runs at 0.089 s per frame, remaining competitive with lightweight baselines and substantially faster than HiSplat (0.510 s).
Additional qualitative results on RealEstate10K are provided in the Appendix.

\begin{table}[!t]
\centering
\caption{
\textbf{Comparison of extrapolated NVS on RealEstate10K.}
We evaluate model performance on RealEstate10K by rendering three novel extrapolated views from two reference views, averaging across all scenes under identical training settings.
}
\resizebox{0.7\linewidth}{!}{
\begin{tabular}{lccc}
\hline
Method & PSNR$\uparrow$ & SSIM$\uparrow$ & LPIPS$\downarrow$ \\
\hline
pixelSplat~\cite{PixelSplat-2024} & 21.76 & 0.779 & 0.217 \\
MVSplat~\cite{MVSplat-2025} & \cellcolor{yellow!20} 21.92 & 0.787 & \cellcolor{yellow!20} 0.199 \\
TranSplat~\cite{aaai2025transplat} & 21.89 &\cellcolor{yellow!20} 0.791 & 0.201 \\
HiSplat~\cite{HiSplat-2024} &\cellcolor{orange!20} 22.01 &\cellcolor{orange!20} 0.794 &\cellcolor{orange!20} 0.191 \\
\textbf{SeeU (Ours)} &\cellcolor{red!20} \textbf{24.33} \textcolor{red}{(+2.32)} &\cellcolor{red!20} \textbf{0.846} \textcolor{red}{(+0.052)} &\cellcolor{red!20} \textbf{0.152} \textcolor{red}{($-$0.039)} \\
\hline
\end{tabular}
}
\label{tab:extrapolation}
\end{table}
\begin{figure}[!t]
    \centering
    \includegraphics[width=\linewidth]{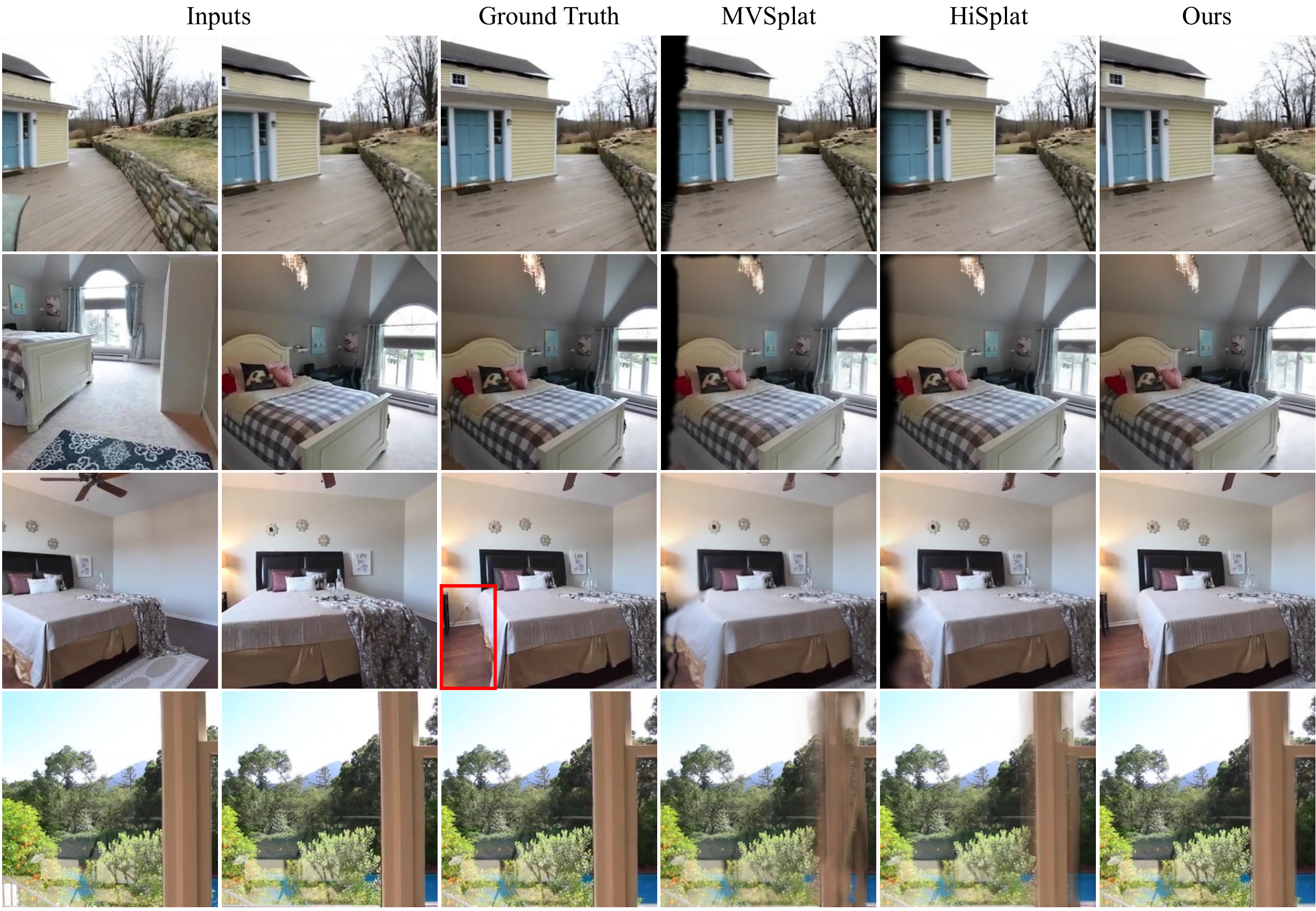}
    \caption{\textbf{Qualitative results for extrapolated NVS on RealEstate10K.}
    Extrapolated target views lie outside the input trajectory where novel pixels lack reference correspondences from input views.
    Baseline methods exhibit voids and distorted geometry within these unseen regions while SeeU reduces missing-geometry artifacts and preserves boundary structures. 
    }
    \label{fig:extrapolation}
    \vspace{-1.5em}
\end{figure}

\subsubsection{Extrapolated Novel View Synthesis}
Extrapolated NVS evaluates target viewpoints outside the reference-view range, where limited input-view support increases reconstruction ambiguity.
For a fair comparison, we retrain all baselines using the same extrapolation-aware training protocol as SeeU, with target views sampled up to 45 frames before and after the reference-view interval.
As summarized in Table~\ref{tab:extrapolation}, SeeU outperforms all baselines, exceeding HiSplat by 2.32 dB in PSNR while reducing LPIPS by 20\%.
Figure~\ref{fig:extrapolation} shows that representative pixel-aligned baselines, MVSplat and HiSplat, exhibit voids and geometric artifacts in weakly constrained regions, whereas SeeU better preserves boundary structures.
Its CEA-conditioned residual refinement incorporates semantic cues beyond pixel-aligned correspondences directly in Gaussian space.

\subsubsection{Zero-Shot Cross-Dataset Evaluation on DTU}
\begin{table}[!t]
\centering
\caption{\textbf{Zero-shot evaluation on DTU.}
All methods are trained on RealEstate10K and evaluated on DTU without fine-tuning.}
\resizebox{0.8\linewidth}{!}{
\begin{tabular}{lccccc}
\hline
Method & PSNR$\uparrow$ & SSIM$\uparrow$ & LPIPS$\downarrow$ & Depth RMSE$\downarrow$ & Overall Chamfer$\downarrow$ \\
\hline
MVSplat~\cite{MVSplat-2025} & 13.94 & 0.473 & 0.385 & 0.847 & 7.37 \\
HiSplat~\cite{HiSplat-2024} & 16.05 & 0.671 & 0.277 & 0.958 & 8.41 \\
\textbf{SeeU (Ours)} & \textbf{16.31} & \textbf{0.704} & \textbf{0.269} & \textbf{0.841} & \textbf{6.75} \\
\hline
\end{tabular}
}
\vspace{-1.5em}
\label{tab:dtu}
\end{table}

\begin{figure}[!t]
    \centering
    \resizebox{0.7\linewidth}{!}{\includegraphics{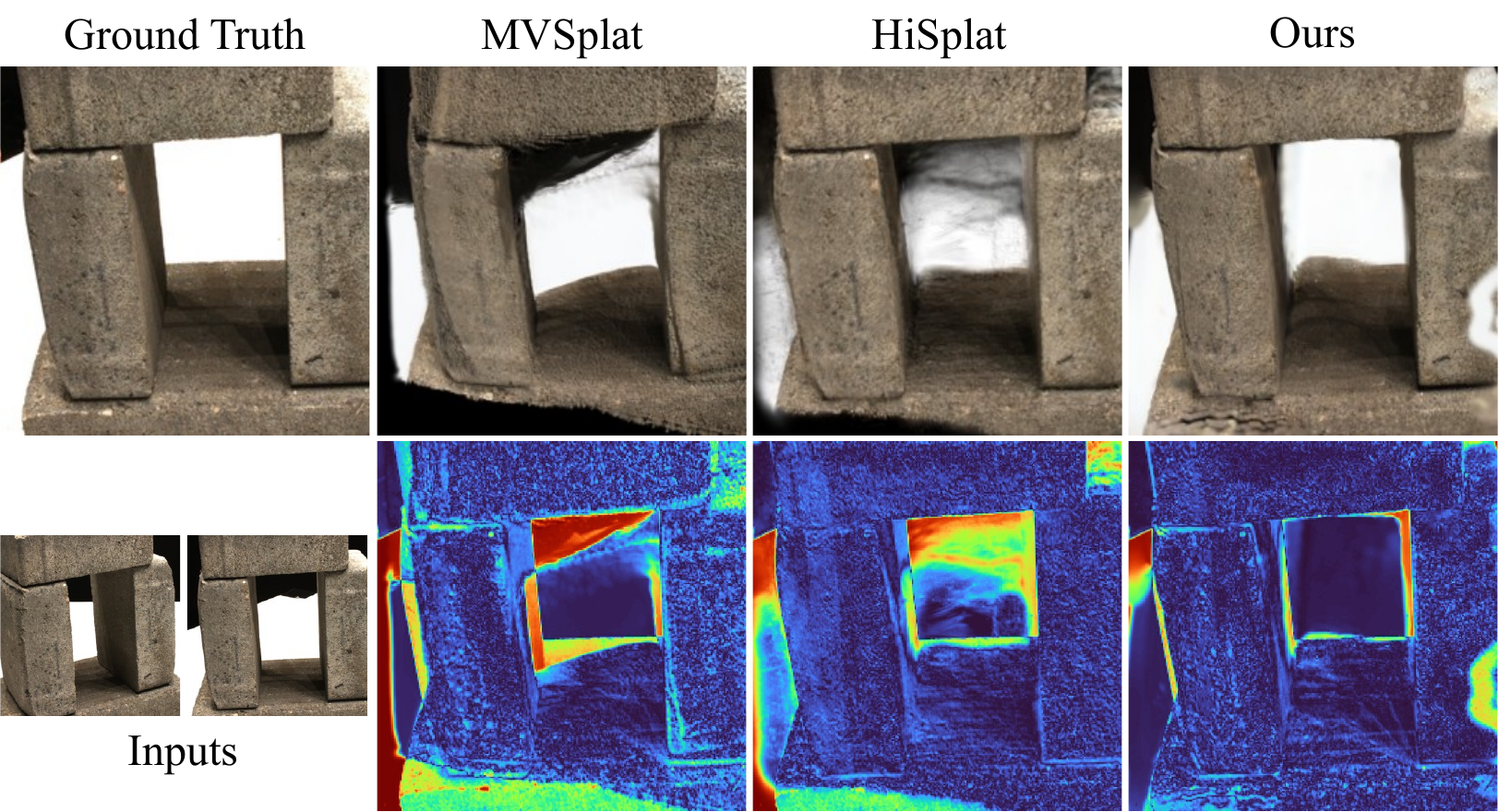}}
    \caption{\textbf{Zero-shot qualitative comparison on DTU.}
    All models are trained on RealEstate10K and evaluated on DTU without fine-tuning.
    The lower-right three panels visualize residuals between the rendered novel views and ground truth.
    }
    \vspace{-1.5em}
    \label{fig:dtu-cross}
\end{figure}

Table~\ref{tab:dtu} shows that SeeU improves rendering fidelity and geometric accuracy simultaneously under cross-dataset transfer.
It achieves better PSNR, SSIM, and LPIPS while also obtaining the lower Depth RMSE (0.841) and Overall Chamfer (6.75).
Notablely, although HiSplat is the strongest rendering baseline, its geometry errors indicate that image-space quality alone does not guarantee accurate 3D reconstruction.
In Figure~\ref{fig:dtu-cross}, SeeU yields smaller residuals and better preserves the boundaries of the partially observed structure.
We attribute this robustness to CEA-conditioned Gaussian refinement, which emphasizes weakly constrained regions using cross-view semantics and matching uncertainty, thereby reducing reliance on precise pixel-aligned depth estimates under domain shift.

\subsection{Ablation Studies}
\begin{table}[t]
\centering
\caption{
\textbf{Ablations of Gaussian refinement.}
Models trained on RealEstate10K with two input views.
\emph{w/o Refinement} renders directly from the initial Gaussians, while \emph{w/ UNet Backbone} replaces the Conditional Gaussian Transformer with a 2D UNet.
Green indicates drops relative to the full model.
}
\resizebox{0.7\linewidth}{!}{
\begin{tabular}{lccc}
\hline
Setup & PSNR$\uparrow$ & SSIM$\uparrow$ & LPIPS$\downarrow$ \\
\hline
w/o Refinement     & 25.47 \textcolor{ForestGreen}{(-2.09)} & 0.849 \textcolor{ForestGreen}{(-0.039)} & 0.149 \textcolor{ForestGreen}{(+0.035)} \\
w/ UNet Backbone   & 26.71 \textcolor{ForestGreen}{(-0.85)} & 0.873 \textcolor{ForestGreen}{(-0.015)} & 0.123 \textcolor{ForestGreen}{(+0.009)} \\
Full Model         & \textbf{27.56}\phantom{ (-0.00)} & \textbf{0.888}\phantom{ (-0.000)} & \textbf{0.114}\phantom{ (+0.000)} \\
\hline
\end{tabular}
}
\label{tab:ablation-refiner}
\end{table}

\begin{figure}[!t]
    \centering
    \includegraphics[width=0.7\linewidth]{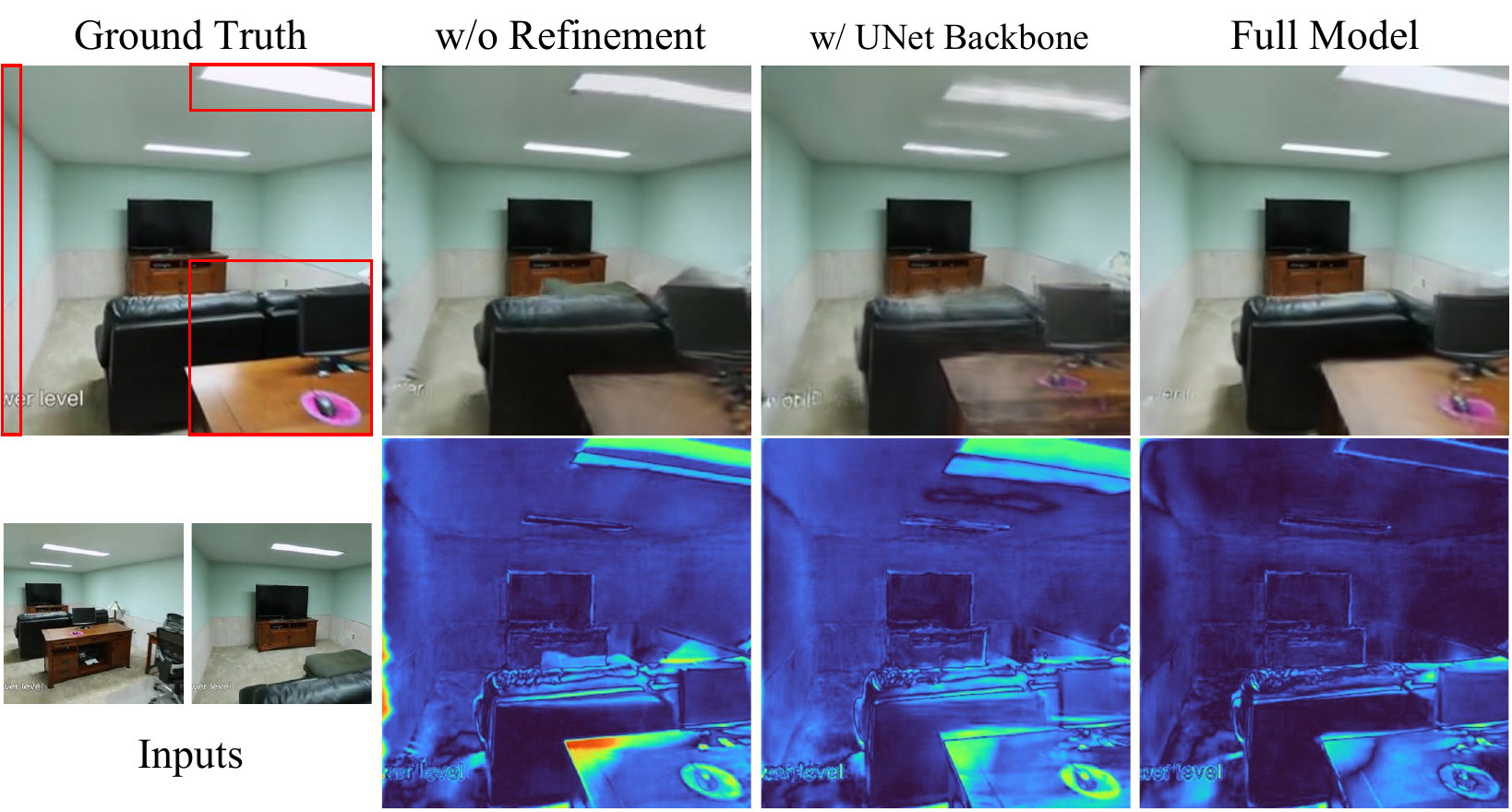}
\caption{
\textbf{Qualitative ablations of Gaussian refinement.}
The second row displays residual maps between rendered images and the ground truth.
Removing Gaussian refinement (\emph{w/o Refinement}) leads to structural distortions and incomplete geometry, while replacing the Conditional Gaussian Transformer with a 2D UNet produces blurry edges.
The full model produces sharper, more complete reconstructions with improved geometric consistency.
}
    \label{fig:ablation-refiner}
\end{figure}

\subsubsection{Ablations on Gaussian Refinement}
We ablate Gaussian refinement from two aspects: (i) removing the refiner (\emph{w/o Refinement}), where the initial Gaussians are directly upsampled for rendering, and (ii) replacing the Conditional Gaussian Transformer with a 2D UNet (\emph{w/ UNet Backbone}).
Quantitative comparisons are summarized in Table~\ref{tab:ablation-refiner}, and qualitative examples under challenging viewpoint shifts are given in Figure~\ref{fig:ablation-refiner}.
Removing refinement leads to pronounced geometric degradation, including collapsed structures and missing surfaces around occlusion boundaries. This confirms that the initial pixel-aligned Gaussians alone are insufficient for reliable reconstruction in sparse-view settings.
Replacing the Conditional Gaussian Transformer with a UNet produces more complete geometry than the no-refinement baseline but introduces blurred contours and texture artifacts (e.g., distorted lighting). We attribute this to convolutional inductive biases that impose rigid 2D grid priors misaligned with the unordered nature of Gaussian primitives.

\begin{table}[!t]
\centering
\caption{
\textbf{Ablations on conditional embeddings.}
Models trained on RealEstate10K with two input views.
CEA outperforms DINOv3, CLIP, and BLIP-2 under identical conditioning interfaces and training settings.
Green indicates drops relative to the best model (CEA).
}
\resizebox{0.7\linewidth}{!}{
\begin{tabular}{lccc}
\hline
Setup & PSNR$\uparrow$ & SSIM$\uparrow$ & LPIPS$\downarrow$ \\
\hline
Class Embedding & 26.13 \textcolor{ForestGreen}{(-1.43)} & 0.859 \textcolor{ForestGreen}{(-0.029)} & 0.139 \textcolor{ForestGreen}{(+0.025)} \\
CLIP Embedding  & 26.41 \textcolor{ForestGreen}{(-1.15)} & 0.867 \textcolor{ForestGreen}{(-0.021)} & 0.126 \textcolor{ForestGreen}{(+0.012)} \\
BLIP Embedding  & 26.54 \textcolor{ForestGreen}{(-1.02)} & 0.870 \textcolor{ForestGreen}{(-0.018)} & 0.125 \textcolor{ForestGreen}{(+0.011)} \\
CEA Embedding   & \textbf{27.56}\phantom{ (-0.00)} & \textbf{0.888}\phantom{ (-0.000)} & \textbf{0.114}\phantom{ (+0.000)} \\
\hline
\end{tabular}
}
\label{tab:ablation-embed}
\end{table}

\begin{figure}[!t]
    \centering
    \includegraphics[width=\linewidth]{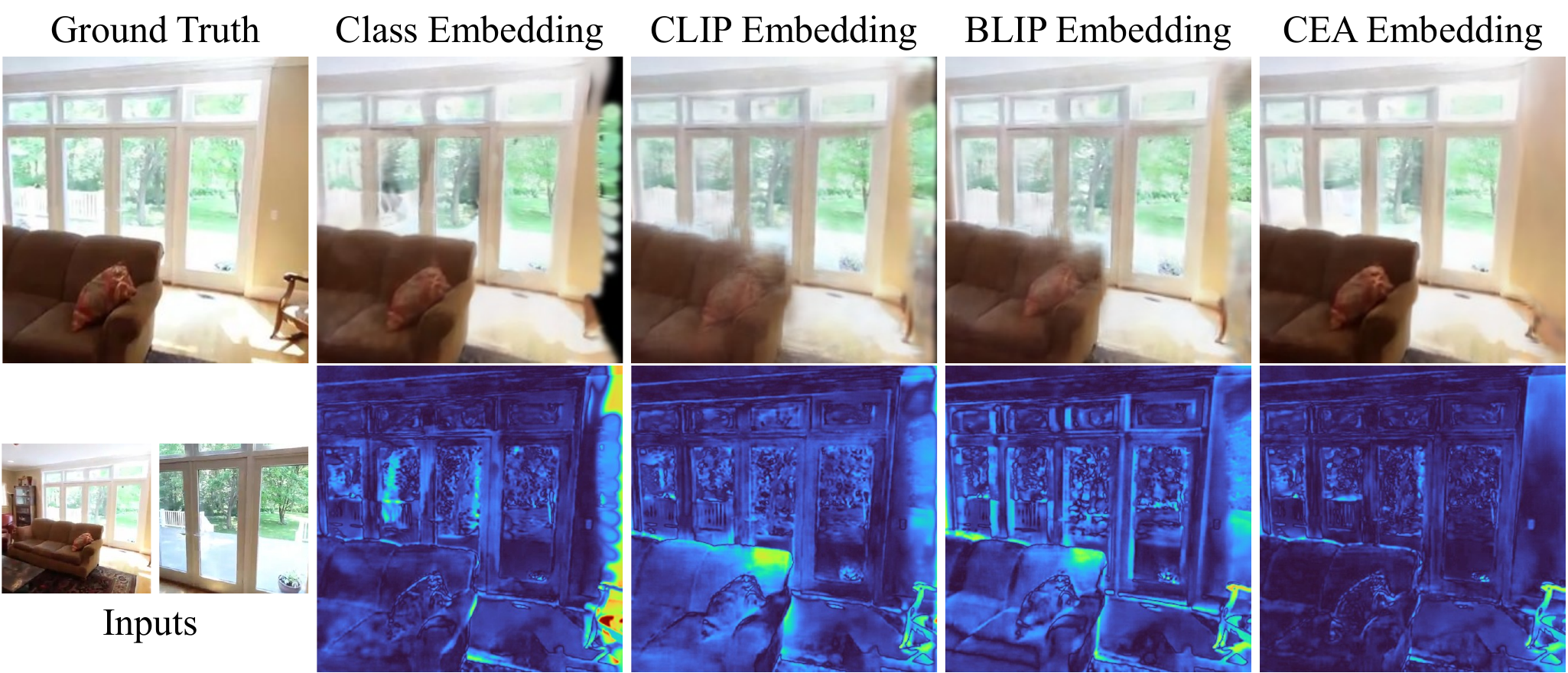}
\caption{
\textbf{Qualitative ablations of various embeddings.} 
The second row displays error maps that indicate differences from the ground truth.
Models using class embeddings exhibit artifacts near boundaries. 
Models conditioned on CLIP and BLIP embeddings exhibit errors along fine structures such as sofa contours. 
The model utilizing the CEA embedding produces sharper edges and exhibits fewer geometric errors.
}
    \label{fig:ablation-embed}
\end{figure}
\subsubsection{Ablations on conditional embeddings}


To assess the role of conditioning, we compare our Cross-view Entropy-Aware (CEA) embedding with three alternatives under identical training and injection configurations: (i) a DINOv3 class embedding, (ii) a CLIP image–text embedding, and (iii) a BLIP-2 pseudo-caption embedding. 
Notably, all pre-trained encoders are kept frozen. 
Their extracted embeddings are linearly projected into a shared 768-dimensional token space and injected via the exact same cross-attention interface, ensuring that the embedding source is the sole variable. 
As summarized in Table~\ref{tab:ablation-embed}, CEA achieves consistent gains across all metrics, improving PSNR by $+1.02$ dB and reducing LPIPS by $\sim 8.8\%$ relative to the next best variant (BLIP).
These gains stem from CEA’s ability to fuse semantics across views while weighting features according to geometric uncertainty. By emphasizing ambiguous or weakly constrained regions, the CEA embedding provides more reliable guidance to the Conditional Gaussian Transformer than single-view or text-derived embeddings, which often capture only global or redundant semantics.
Qualitative comparisons in Figure~\ref{fig:ablation-embed} corroborate these trends. Embeddings derived from individual views (DINOv3) tend to overlook boundary information and may introduce spurious artifacts (e.g., tree-like structures outside the window). 
CLIP and BLIP embeddings better capture salient global content but still struggle with fine structures such as sofa contours.
This could be attributed to their sensitivity to multi-view overlap and redundancy, which diminishes the attention to unique objects appearing only once. 
In contrast, CEA recovers sharper edges and more complete geometry, demonstrating its capacity to highlight uncertain regions while suppressing redundancy.

\subsubsection{Extrapolation Training Range}
\begin{table}[!t]
\centering
\caption{
\textbf{Ablations on extrapolation training range.}
We vary the sampling range of extrapolated target views during training on RealEstate10K with two input views.
\emph{w/ no ext.} disables extrapolated-view sampling, whereas \emph{ext. $N$ frames} samples targets within $N$ frames before and after the input-view range.
Green indicates drops relative to the 45-frame setting.
}
\resizebox{0.7\linewidth}{!}{
\begin{tabular}{lccc}
\hline
Variant & PSNR$\uparrow$ & SSIM$\uparrow$ & LPIPS$\downarrow$ \\
\hline
SeeU w/ no ext.       & 22.78 \textcolor{ForestGreen}{(-1.55)} & 0.804 \textcolor{ForestGreen}{(-0.042)} & 0.178 \textcolor{ForestGreen}{(+0.026)} \\
SeeU ext. 90 frames   & 23.41 \textcolor{ForestGreen}{(-0.92)} & 0.820 \textcolor{ForestGreen}{(-0.026)} & 0.169 \textcolor{ForestGreen}{(+0.017)} \\
SeeU ext. 45 frames   & \textbf{24.33}\phantom{ (-0.00)} & \textbf{0.846}\phantom{ (-0.000)} & \textbf{0.152}\phantom{ (+0.000)} \\
\hline
\end{tabular}
}
\label{tab:ablation-ext-range}
\end{table}

We vary the range of extrapolated target views sampled during RealEstate10K training.
Table~\ref{tab:ablation-ext-range} shows that removing extrapolated targets reduces rendering performance, demonstrating the importance of extrapolation-aware supervision.
The 90-frame setting still improves over no extrapolation, but remains consistently below the 45-frame setting.
At very large offsets, target views may expose substantially more content with limited support from the inputs, increasing reconstruction ambiguity and weakening the supervision signal.
A moderate range instead presents challenging out-of-interval views while retaining sufficient visual evidence to associate semantic cues with scene geometry.
This balance improves rendering fidelity and structural preservation, indicating that overly aggressive extrapolation does not necessarily yield better generalization.

\section{Conclusion}
In this work, we propose SeeU, a novel G-3DGS framework for novel view synthesis from sparse-view inputs. 
Unlike prior methods that rely on pixel-aligned Gaussian estimation, our approach uses a CEA-conditioned Conditional Gaussian Transformer to apply residual updates directly in Gaussian space, improving geometry in partially observed or uncertain scenes.
To enhance both global semantic coherence and fine-grained detail awareness, we introduce a Cross-view Entropy-Aware embedding that provides semantically rich guidance for Gaussian refinement.
Extensive experiments on multiple datasets demonstrate that SeeU consistently enhances reconstruction fidelity, particularly in occluded and unseen regions.
These results highlight the potential of semantic-conditioned refinement in Gaussian space for generalizable 3D reconstruction.



%
%
\bibliographystyle{splncs04}
\bibliography{main}

\clearpage
\appendix
\renewcommand{\theHsection}{appendix.\Alph{section}}
\renewcommand{\theHsubsection}{appendix.\Alph{section}.\arabic{subsection}}

\section{Implementation Details}\label{appendix:implydetail}
We implement SeeU in PyTorch.
Unless otherwise specified, all input images are resized to $256\times256$, and two sparse input views are used.
With a downsampling factor of $s=4$, the latent resolution is $64\times64$.
For semantic feature extraction, we use a frozen ViT-Base encoder~\cite{iclr2020vit} pretrained with DINOv3~\cite{simeoni2025dinov3}.
The Conditional Gaussian Transformer follows the TinyDiT architecture~\cite{cvpr2025tinyfusion} and comprises four layers with a hidden size of 256, four attention heads, and an MLP ratio of 2.
We train the model for 300{,}000 iterations on one NVIDIA H800 GPU using AdamW~\cite{iclr2019adamw} with a batch size of 8.

\section{Additional Discussion}
\subsection{Zero-Shot Evaluation on ACID}\label{appendix:acid-cross}
\begin{figure}[!th]
    \centering
    \resizebox{0.7\linewidth}{!}{\includegraphics{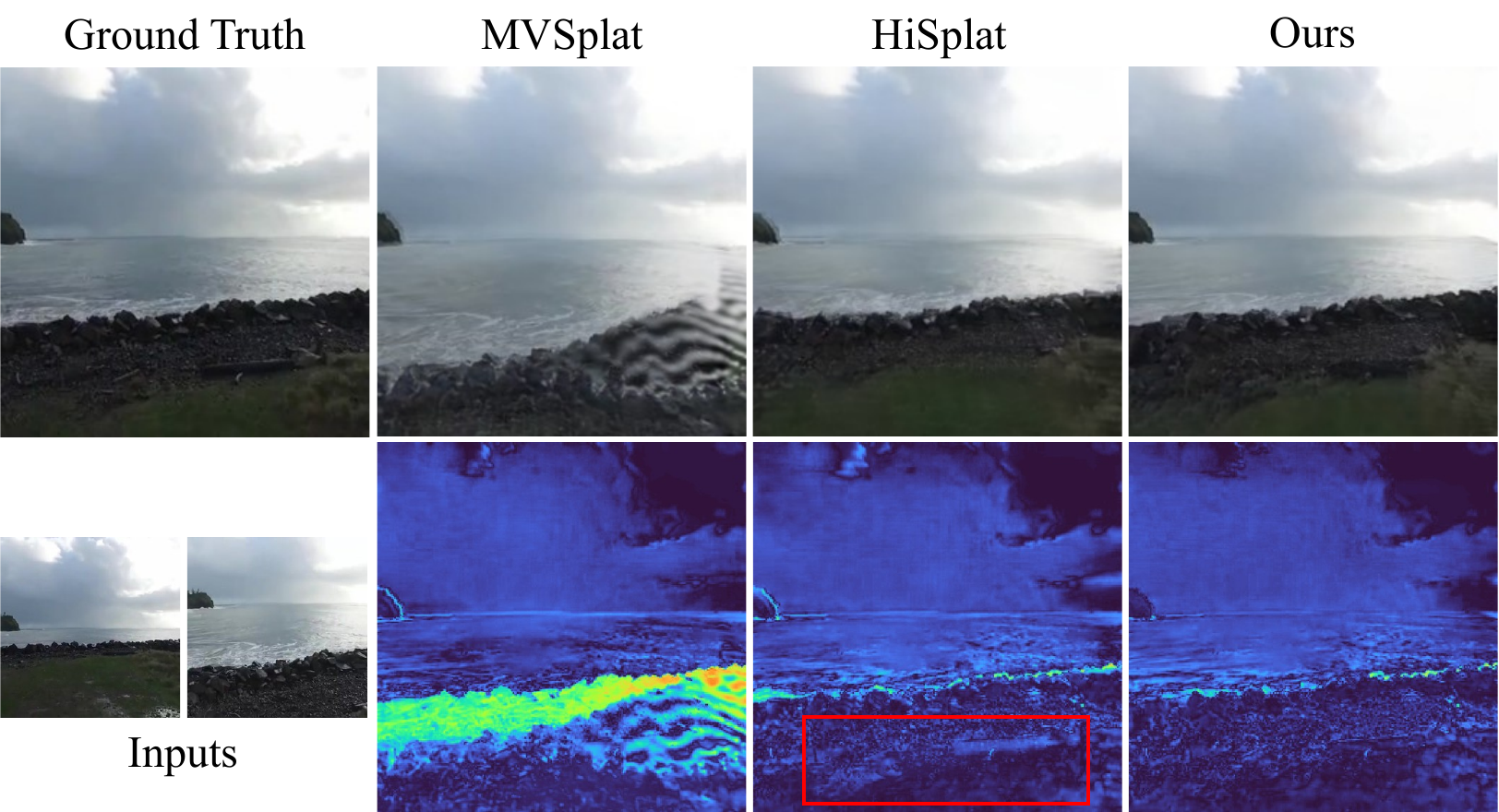}}
    \caption{\textbf{Zero-shot qualitative comparison on ACID.}
    All models are trained on RealEstate10K and evaluated on ACID without fine-tuning.}
    \label{fig:acid-cross}
\end{figure}

We additionally evaluate the RealEstate10K-trained models on ACID without fine-tuning.
As shown in Table~\ref{tab:acid-cross}, SeeU achieves the best PSNR and SSIM, improving over HiSplat by 0.04 dB and 0.003, respectively, while HiSplat retains a marginal 0.001 advantage in LPIPS.
Figure~\ref{fig:acid-cross} provides the corresponding qualitative comparison under this domain shift.

\begin{table}[!th]
\centering
\caption{\textbf{Zero-shot evaluation on ACID.}
All methods are trained on RealEstate10K and evaluated on ACID without fine-tuning.}
\resizebox{0.5\linewidth}{!}{
\begin{tabular}{lccc}
\hline
Method & PSNR$\uparrow$ & SSIM$\uparrow$ & LPIPS$\downarrow$ \\
\hline
pixelSplat~\cite{PixelSplat-2024} & 27.64 & 0.830 & 0.160 \\
MVSplat~\cite{MVSplat-2025} & 28.15 & 0.841 & 0.147 \\
TranSplat~\cite{aaai2025transplat} & 28.17 & 0.842 & 0.146 \\
HiSplat~\cite{HiSplat-2024} & 28.66 & 0.850 & \textbf{0.137} \\
\textbf{SeeU (Ours)} & \textbf{28.70} & \textbf{0.853} & 0.138 \\
\hline
\end{tabular}
}
\label{tab:acid-cross}
\end{table}

\begin{table}[t]
    \centering
    \caption{\textbf{Quantitative comparison of 3-view cross-dataset generalization.}}
    \resizebox{0.6\linewidth}{!}{
    \begin{tabular}{lccc}
        \hline
        Methods & PSNR $\uparrow$ & SSIM $\uparrow$ & LPIPS $\downarrow$ \\
        \hline
        pixelSplat~\cite{PixelSplat-2024} & 12.52 & 0.367 & 0.585 \\
        MVSplat~\cite{MVSplat-2025} & 14.30 & 0.508 & 0.371 \\
        HiSplat~\cite{HiSplat-2024} & 16.34 & 0.674 & \textbf{0.286} \\
        \textbf{SeeU (Ours)} & \textbf{16.44} {(+0.1)} & \textbf{0.729} {(+0.055)} & 0.298 {(+0.012)} \\
        \hline
    \end{tabular}
    }
    \label{tab:3view}
\end{table}

\subsection{Generalization to More Input Views}\label{appendix:more-view}
To assess the generalizability of our model under varying numbers of sparse input views, we conduct a zero-shot evaluation by directly applying the model trained with 2-view inputs in RealEstate10K to a 3-view input setting on DTU.
As reported in Table~\ref{tab:3view}, our method demonstrates robust performance when evaluated with more input views.

\begin{figure*}[!t]
    \centering
    \includegraphics[width=\linewidth]{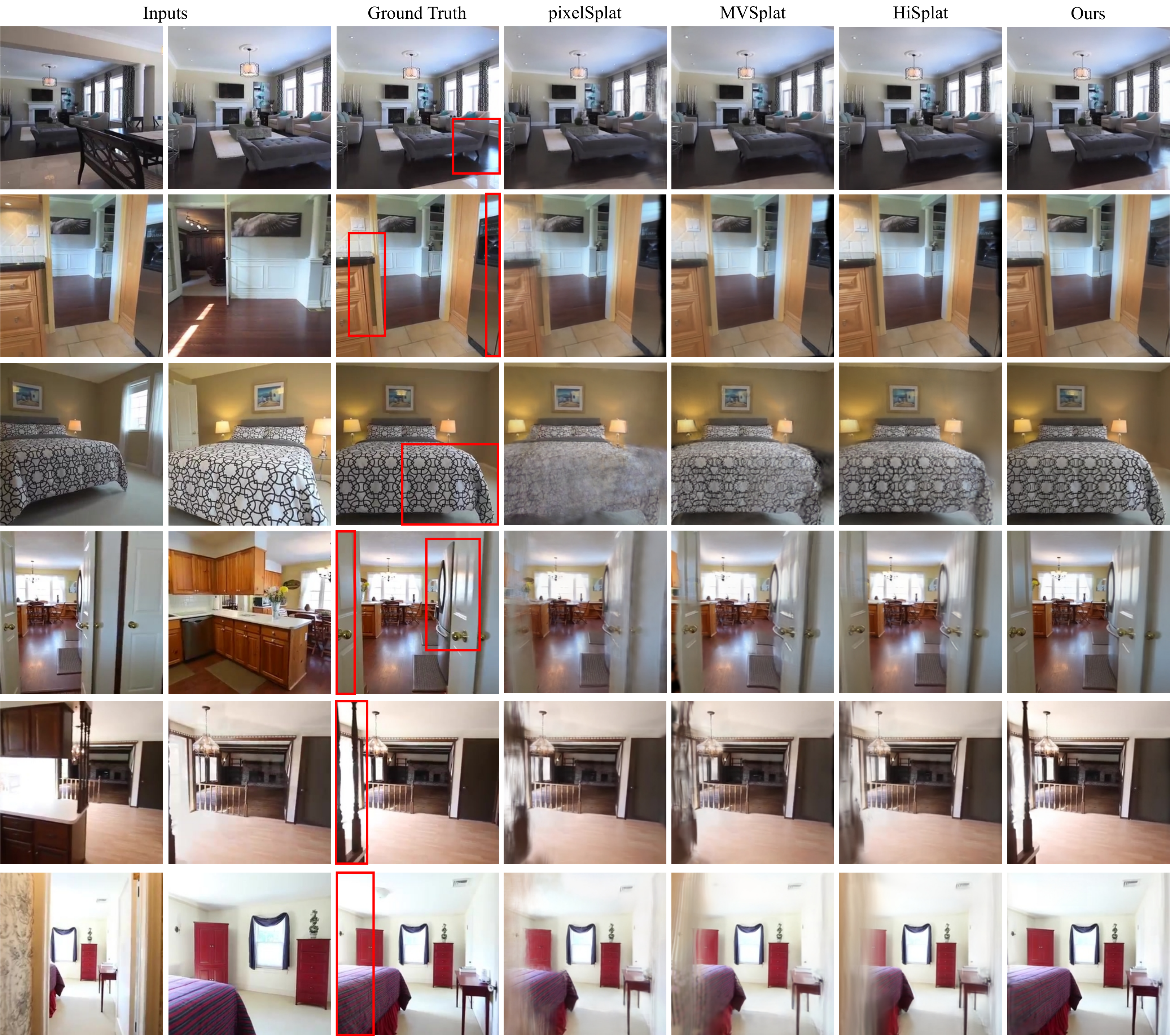}
    \caption{\textbf{More comparisons on RealEstate10K.} Qualitative comparison of novel view synthesis results across different methods.
Red boxes highlight challenging regions where baseline methods exhibit artifacts or texture loss.
In contrast, SeeU produces sharper and more structurally consistent renderings, refining object contours and details (e.g., furniture edges, patterned bedspreads).}
    \label{fig:comparison-appendix}
    \vspace{-1.2em}
\end{figure*}

\subsection{More Visual Comparisons}\label{appendix:compare}
We provide additional qualitative comparisons on the RealEstate10K dataset, as shown in Figure~\ref{fig:comparison-appendix}. PixelSplat and MVSplat frequently exhibit severe blurring and geometric distortions, while HiSplat produces sharper outputs but still fails to recover boundaries and occluded regions.
In contrast, SeeU consistently generates sharper novel views, preserving furniture outlines and textile patterns.
\end{document}